\documentclass[journal,twoside,web]{ieeecolor}
\usepackage{generic}
\usepackage{amsmath,amssymb,amsfonts}
\usepackage{algorithmic}
\usepackage{graphicx}
\usepackage{algorithm,algorithmic}
\usepackage{hyperref}
\hypersetup{hidelinks=true}
\usepackage{textcomp}
\def\BibTeX{{\rm B\kern-.05em{\sc i\kern-.025em b}\kern-.08em
    T\kern-.1667em\lower.7ex\hbox{E}\kern-.125emX}}
\usepackage[maxbibnames=20,sorting=nty,doi=false,isbn=false,url=false,eprint=false,maxcitenames=20]{biblatex}
\begin{document}

\title{Grounding Chest X-Ray Visual Question Answering with Generated Radiology Reports}

\author{Francesco Dalla Serra, Patrick Schrempf, 
Chaoyang Wang, Zaiqiao Meng, \\Fani Deligianni, and Alison Q. O'Neil
\thanks{Francesco Dalla Serra is with Canon Medical Research Europe, Edinburgh, EH65NP, UK (e-mail: francesco.dallaserra@mre.medical.canon).}
\thanks{Patrick Schrempf is with Canon Medical Research Europe, Edinburgh, EH65NP, UK.}
\thanks{Chaoyang Wang is with Canon Medical Research Europe, Edinburgh, EH65NP, UK.}
\thanks{Zaiqiao Meng is with School of Computing Science, University of Glasgow, Glasgow, G128QQ, UK.}
\thanks{Fani Deligianni is with School of Computing Science, University of Glasgow, Glasgow, G128QQ, UK.}
\thanks{Alison Q. O'Neil is with Canon Medical Research Europe, Edinburgh, EH65NP, UK.}}

\maketitle

\begin{abstract}
We present a novel approach to Chest X-ray (CXR) Visual Question Answering (VQA), addressing both single-image image-difference questions. Single-image questions focus on abnormalities within a specific CXR (\emph{``What abnormalities are seen in image X?''}), while image-difference questions compare two longitudinal CXRs acquired at different time points (\emph{``What are the differences between image X and Y?''}). We further explore how the integration of radiology reports can enhance the performance of VQA models. While previous approaches have demonstrated the utility of radiology reports during the pre-training phase, we extend this idea by showing that the reports can also be leveraged as additional input to improve the VQA model's predicted answers. First, we propose a unified method that handles both types of questions and auto-regressively generates the answers. For single-image questions, the model is provided with a single CXR. For image-difference questions, the model is provided with two CXRs from the same patient, captured at different time points, enabling the model to detect and describe temporal changes. Taking inspiration from 'Chain-of-Thought reasoning', we demonstrate that performance on the CXR VQA task can be improved by grounding the answer generator module with a radiology report predicted for the same CXR. In our approach, the VQA model is divided into two steps: i) Report Generation (RG) and ii) Answer Generation (AG). Our results demonstrate that incorporating predicted radiology reports as evidence to the AG model enhances performance on both single-image and image-difference questions, achieving state-of-the-art results on the Medical-Diff-VQA dataset.
\end{abstract}


\section{Introduction}

\begin{figure*}[th!]
    \centering
    \includegraphics[width=\textwidth]{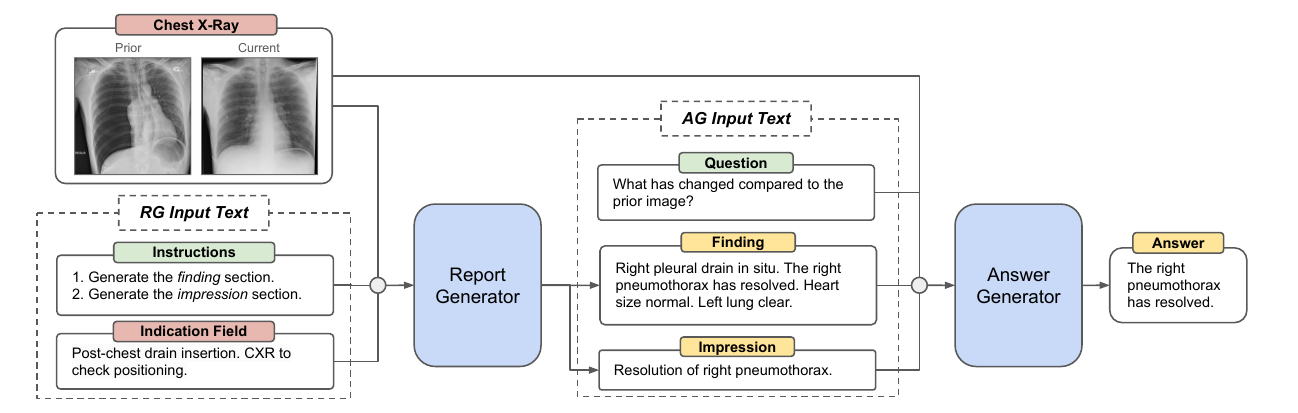}
    \caption{Overview of the Report Generator–Answer Generator (RG-AG) pipeline: (1) the Report Generator first produces a radiology report based on the given Chest X-ray (or a pair of images in the case of a follow-up study), along with the instruction and the indication field. The report consists of the `finding' and `impression' sections, which are generated independently based on the specific instructions received by the RG module. (2) The Answer Generator then utilises this predicted report as additional contextual information to generate a more accurate and interpretable response to the input question. Red tags denote clinical input data, green tags indicate the prompts provided to each module, and yellow tags represent the output generated by each module.}
    \label{fig:rg_vqa}
\end{figure*}

Visual Question Answering (VQA) refers to the task of answering questions about the contents of an image. Multiple potential roles have been suggested for medical VQA, such as a `second opinion' in image interpretation, and as an assistant in answering imaging-based queries of clinicians to improve the radiology workflow \cite{lin2023medical}. 

The growing volume of medical imaging studies \cite{smith2008rising}, driven in part by an ageing population, continues to place a heavy burden on radiologists. This strain, compounded by institutional limitations and a global shortage of radiologists \cite{rimmer2017radiologist, cao2023current}, can lead to diagnostic delays that adversely affect patient care \cite{Care2018}. AI diagnostic assistance has the ability to improve the clinical accuracy of radiologists of CXR reporting \cite{yu2024heterogeneity}, therefore, a complete radiology VQA that can understand free-form questions and produce reliable answers has attracted a lot of interest.  

In this paper, we focus on VQA applied to Chest X-Ray (CXR) radiology. CXR images are 2D projections through various overlapping structures within the thoracic cavity, such as the ribs, heart, and lungs. This makes it challenging to isolate and identify abnormalities in specific structures. Furthermore, the differences in the appearance of normal and abnormal findings are often subtle and can be difficult to detect. The multimodal nature of VQA adds another layer of complexity, as it requires the system to interpret visual information in the context of a textual query. While this presents a powerful tool for advancing medical diagnostics, it also introduces significant challenges.

VQA can be open-ended, requiring the generation of free-form text answers; closed-ended, requiring the generation of short specific responses; or multiple choice, where the model selects the correct answer from a set of provided options. Additionally, questions may refer to a single image or comparisons between two images to identify differences. Questions based on a single image (\textit{e.g.}, closed-ended question: ``Is there any sign of pneumonia in the given scan?'' [Answer: Yes/No]) can facilitate more efficient decision-making by clinical teams, prior to the formal report issued by a radiologist. Questions that compare scans of the same patient taken at different time points (\textit{e.g.}, open-ended questions: ``What has changed compared to the prior scan?'',  ``Has the effusion resolved as expected?'',  ``Is the rate of change of the nodule concerning for a malignant lesion?'') are essential for monitoring disease progression or treatment response. We refer to the second as \textit{image-difference question answering}. This is particularly relevant in the medical domain, where radiologists often compare scans from different timepoints to assess the progression of findings.

To effectively handle diverse VQA scenarios, we propose a flexible vision-language model that processes dual visual and textual information inputs, tailoring its input configuration based on the specific VQA task we want to perform. Our approach adopts \textit{anatomy-finding anatomical tokens} \cite{dallaserra2023atlastokens} and a \textit{longitudinal projection module} \cite{serra-etal-2023-controllable}. Moreover, we study the effect of grounding the answer generation module using a radiology report predicted from the same image. Our approach is inspired by Chain-of-Thought (CoT) reasoning, as demonstrated in language-only QA tasks \cite{wei2022chain, kojima2022large}, and more recently in vision-language models \cite{zhang2024multimodal}. In these settings, having the model explicitly generate a reasoning process before producing an answer has been shown to improve inference quality.

We propose that providing additional supporting evidence to a VQA model could improve answer prediction. Specifically, we hypothesise that generating a comprehensive description of a CXR’s appearance in the form of a detailed radiology report---based on the same scan the question refers to---and grounding the VQA model with this predicted report could enhance both the accuracy and reliability of the generated answer. Our focus is on generating the two main sections of a radiology report: the `findings' section, which offers a detailed account of the scan’s visual characteristics, and the `impression' section, where radiologists synthesise the findings to form a cohesive clinical interpretation.

We further incorporate successful strategies from previous works on the diff-VQA task, namely anatomical feature representations \cite{hu2023expert} and a pre-training strategy based on radiology report generation \cite{cho2024pretraining}. To the best of our knowledge, this is the first study to examine the impact of grounding radiology reports for medical VQA and to provide evidence of its effectiveness on both single-image and image-difference VQA.

In summary, our contributions consist of the following.
\begin{enumerate}
\item Propose a unified approach---the Report Generator–Answer Generator (RG-AG) pipeline---for addressing both single-image and image-difference CXR VQA tasks.
\item Show the effectiveness of grounding the answer generation process using the corresponding radiology report to improve the quality and accuracy of the response.
\item Achieve state-of-the-art performance on the publicly available Medical-Diff-VQA dataset \cite{hu2023medical, hu2023expert}, with the most significant improvements observed on image-difference questions and single-image open-ended questions.
\end{enumerate}

\section{Related Works}

\subsection{Medical Visual Question Answering}

The limited diversity of answers in medical VQA datasets \cite{slake, RAD-VQA} has often led researchers to approach medical VQA as a classification task \cite{nguyen2019overcoming, mmbert, do2021multiple}. However, treating VQA as a classification task limits the solution to a predetermined set of answers. Generative approaches \cite{cgmvqa, sharma2021medfusenet, van2023open} have been enabled by the availability of open-access datasets \cite{zhang2023pmc, humedical} and the rise of generative large language models \cite{touvron2023llama, OpenAI2023GPT4TR}. Treating VQA as a text-generation task naturally yields more detailed and wide-ranging responses expressed in the form of one or multiple sentences. Some authors \cite{cgmvqa, sharma2021medfusenet} have proposed medical VQA methods that use two parallel heads, allowing the model to perform either classification or answer generation. These heads are trained separately depending on the answer type, whether closed-ended or open-ended. More recently, in \cite{van2023open} the authors proposed a VLM that integrates a vision encoder with GPT-2 \cite{radford2019language}, a large language model pre-trained on a general corpus, demonstrating the model's effectiveness by fine-tuning it on three medical VQA datasets \cite{slake, pathvqa, huang2022ovqa}.

\subsection{Medical Image-Difference Question Answering}

\textit{Image-difference question answering}, henceforth referred to as \textit{diff-VQA}, is the task where questions refer to the differences between two or more images. This task has received limited attention in both the general \cite{Qiu_2021_ICCV, yao2022image} and medical \cite{hu2023expert, cho2024pretraining} domains. In the medical domain, this is primarily due to the lack of a suitable dataset, until the creation of Medical-Diff-VQA \cite{hu2023expert}---a CXR VQA dataset designed to include such questions. \cite{hu2023expert} propose a method for diff-QA that utilises anatomical features and a multi-relationship image-difference graph feature representation learning method to extract image-difference features. To the best of our knowledge, only one other work \cite{cho2024pretraining} has addressed diff-VQA. The authors adopt a pre-trained VLM \cite{wang2022ofa} and propose a multi-stage pre-training pipeline (PLURAL), where the VLM is first fine-tuned on the report generation task and subsequently fine-tuned on the VQA task. However, this approach does not leverage the generated radiology reports during the VQA task.

\subsection{Grounding CXR-VQA with Radiology Reports}

To the best of our knowledge, the use of predicted radiology reports to enhance VQA performance has not been explored in the literature. The most relevant work is by \cite{wang2023chatcad}, who proposed a method that integrates the output of Computer-Aided Diagnosis (CAD) networks with a Large Language Model (LLM) to leverage the LLM's medical domain knowledge and logical reasoning. However, their approach focuses on leveraging LLMs to improve the interactivity of a CAD network. Conversely, we focus on demonstrating how CAD networks can enhance the VQA performance of a VLM.

\section{Method}

We present an overview of our \textit{Report Generator}-\textit{Answer Generator} (RG-AG) model in Figure \ref{fig:rg_vqa}.

\subsection{Visual Anatomical Tokens Extractor}

We extract visual CXR features in the form of finding-aware anatomical tokens \cite{dallaserra2023atlastokens} i.e. vector representations corresponding to a predefined set of anatomical regions in the CXR.

To generate tokens, we train a Faster R-CNN model \cite{ren2015faster} to perform two tasks: (1) \textit{anatomical region localisation} -- detecting the bounding box of $N=36$ anatomical regions; and (2) \textit{finding detection} -- determining the presence or absence of 71 findings within each region. Tokens are extracted for each CXR by selecting the bounding box representation with the highest confidence score for each anatomical location and extracting the corresponding feature vectors from the Region of Interest pooling layer of the Faster R-CNN. This results in $N$ vectors $V=\{\Vec{v}_n\}_{n=1}^N$ with $\Vec{v}_n\in\mathbb{R}^{d}$ and $d$=1024. If an anatomical region $i$ is not detected in a CXR, the corresponding token $\Vec{v}_i$ is a zero vector. For more details about the training and the model architecture, we refer the reader to \cite{dallaserra2023atlastokens}. These tokens are input to both the RG and VQA modules. 

\subsection{Vision-Language Model Architecture}

\begin{figure*}
    \centering
    \includegraphics[width=\textwidth]{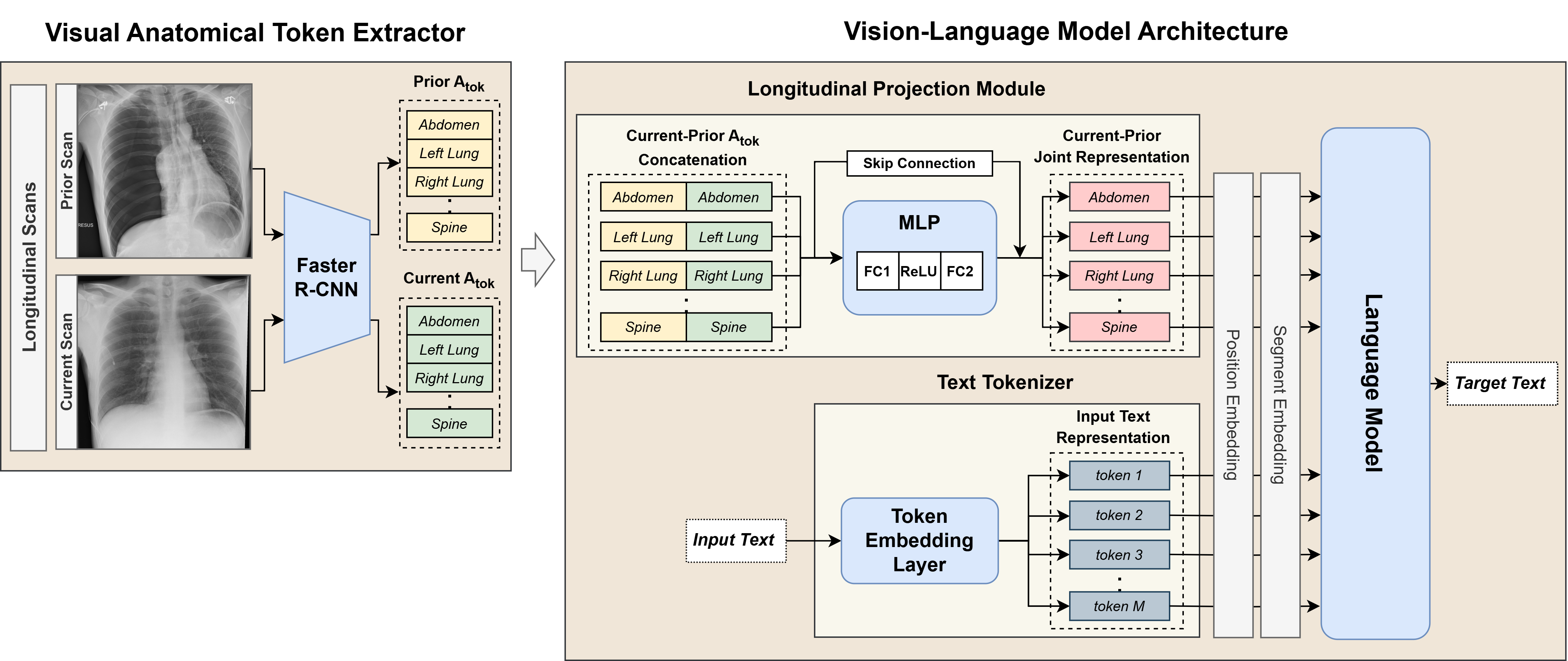}
    \caption{The model architecture of the Report Generator and Answer Generator. This diagram illustrates on the left the \textit{Visual Anatomical Token Extractor} -- responsible for extracting the visual tokens from CXRs. This component is trained independently of the Vision-Language Model. On the right, the \textit{Vision-Language Model} architecture is responsible for generating the radiology report or the answer. The diagram shows how visual inputs (\textit{i.e.}, anatomical tokens $A_{tok}$) are aligned, concatenated and projected into a joint representation via the \textit{Longitudinal Projection Module}. This representation is then combined with tokenised text inputs, which the language model processes to generate the target text. For the Report Generator, the input text is an instruction requesting the generation of a specific report section (finding or impression), with the target text being that section. For the Answer Generator, the input text is the concatenation of the question and the predicted report (finding + impression), and the target text is the answer.}
    \label{fig:rg_vqa_architecture}
\end{figure*}

The RG and AG use the same VLM architecture, composed of a \textit{Longitudinal Projection Module} (LPM) and a Transformer-based Language Model (LM). This architecture has a total of 68M trainable parameters. The VLM is defined as a function $f$ which takes visual features ($V$) and text ($T$) as inputs, and generates output text ($Y$):

\begin{equation}\label{eq:vlm}
    Y = f(V,T),
\end{equation}

\noindent
which is achieved in an autoregressive manner. The text inputs and outputs vary based on the specific task (RG or AG). The VLM architecture of the RG and AG models are shown in Figure \ref{fig:rg_vqa_architecture}.

\subsubsection{Longitudinal Projection Module}

The LPM is responsible for projecting the current and prior CXR scans into a joint representation. The input visual components of the LPM correspond to the finding-aware anatomical tokens from the current and for image-difference types of questions, the prior scan is also provided. These are denoted as $V_{c} = \{\Vec{v}_{c,n}\}_{n=1}^N$ and $V_{p} = \{\Vec{v}_{p,n}\}_{n=1}^N$, respectively. Whenever we do not intend to use the prior scan as input, we set $V_{p} = \{\Vec{0}\}_{n=1}^N$. 

We first align and concatenate each pair of anatomical tokens for the current and prior scans assigned to anatomical region $n$ i.e., ($[\Vec{v}_{c,n}, \Vec{v}_{p,n}]$). We then pass the sequence of $N$ pairs into the Multi-Layer Perceptron (MLP), which consists of a stack of a Fully-Connected layer (FC1), a ReLU function and another Fully-Connected layer (FC2). The output of the MLP is added to its input via a residual connection, yielding:

\begin{equation}
    \Vec{v}_{joint, n} = MLP([\Vec{v}_{c,n}, \Vec{v}_{p,n}]) + [\Vec{v}_{c,n}, \Vec{v}_{p,n}]
\end{equation}

\noindent
This formulation preserves the original input information while allowing the MLP to refine the representation. We refer to the output of the LPM as the current-prior joint representation $V_{joint} = \{\Vec{v}_{joint,n}\}_{n=1}^N$.

\subsubsection{Language Model}

The LM consists of a vanilla encoder-decoder Transformer \cite{vaswani2017attention}. Both the encoder and the decoder are composed of 3 attention layers with 8 heads and 512 hidden units. The LM takes as input the current-prior joint representation $V_{joint}$ and the tokenised input text embeddings $T$ and generates the target text $Y$:

\begin{equation}
    Y = LM(V_{joint}, T).
\end{equation}

\noindent
The input visual $V_{joint}$ and text embeddings $T$ are concatenated, and summed to the \textit{position embeddings} -- which establish the position of each token within the input sequence. The \textit{segment embedding} is used to enable the model to distinguish between the modality of each set of input tokens: vision \textit{vs.} text. 

While the RG and AG models share the same architecture design, the input and output components are different, as described in the following sections.

\subsection{Report Generator}
\label{sec:rep_gen}

The Report Generator (RG) is responsible for generating a report describing the target (current) CXR. Specifically, it generates the two main sections of the radiology report---the `findings' and `impression' sections---independently, using distinct prompts for each. The RG model consists of the LPM and the LM. The visual input comprises the current ($V_{c}$) and prior anatomical tokens ($V_{p}$), when available. The target text $Y$ is \textit{either} the finding ($F$) \textit{or} the impression ($I$) section of a radiology report, and the text input $T$ comprises not only the indication field ($\{ind\}$) but also an instruction specifying the section to be generated: (1) $Inst_f=$\textit{``generate the finding section''} or (2) $Inst_i=$\textit{``generate the impression section''}. Following Eq. \ref{eq:vlm}, the RG can be defined as a function $f_{RG}$:

\begin{equation}\label{eq:rr}
\begin{split}
    F &= f_{RG}(V=[V_{c}, V_{p}], T=``\texttt{[RG]}\{ind\}\texttt{[Q]} Inst_f \text{''})\\
    I &= f_{RG}(V=[V_{c}, V_{p}], T=``\texttt{[RG]}\{ind\}\texttt{[Q]} Inst_i \text{''})
\end{split}
\end{equation}

\noindent
where \texttt{[RG]} corresponds to a special token indicating the report generation task, and \texttt{[Q]} is the special token used in front of the instruction.

\subsection{Answer Generator}\label{sec:vqa_model}

The Answer Generator (AG) is responsible for performing VQA. Similar to the RG, it consists of the LPM and the LM. Following \cite{cho2024pretraining}, we initialise the AG model using the RG weights, i.e. the RG task effectively acts as pre-training. The AG model can be defined as a function $f_{AG}(V, T)$, whose input visual features ($V$) and input text ($T$) vary depending on the question type. 

When the question asks about the difference in appearance between two scans, the AG takes as input two sets of anatomical tokens (\textit{current} and \textit{prior}). Otherwise, the AG takes only one set of anatomical tokens (considered to be \textit{current}) and sets $V_{p} = \varnothing =  \{\Vec{0}\}_{n=1}^N$. This results in the visual component being:

\begin{align}
    V = 
    \begin{cases} 
        (V_{c}, V_{p}),   &\mbox{if diff-VQA}; \\ 
        (V_{c}, \varnothing), &\mbox{otherwise}.
    \end{cases}
\end{align}

Additionally, the input text $T$ varies depending on whether the question $\{q\}$ is open-ended or closed (multiple choice):

\begin{align}
    T = 
    \begin{cases} 
        ``\texttt{[OE\_VQA]} \{rr\} \texttt{[Q]} \{q\} \text{''} \quad \mbox{if \textit{\{q\}} is open-ended}, \\[6pt]
        ``\texttt{[MC\_VQA]}  \{rr\}  \texttt{[Q]} \{q\} \texttt{[MC]} \{a_1\}\dots \texttt{[MC]} \{a_M\} \text{''} \\ 
        \qquad \mbox{if \textit{\{q\}} is multiple choice}.
    \end{cases}
\end{align}

\noindent
where $\{rr\}$ refers to the predicted radiology report for the given scan, obtained by concatenating the findings ($F$) and impression sections ($I$) from Eq. \ref{eq:rr}. We define \texttt{[OE\_VQA]} and \texttt{[MC\_VQA]} as special tokens, used to specify the task as open-ended VQA or multiple-choice VQA to the model. \texttt{[MC]} is a special token placed before each possible answer $\{a_j\}$, from which the model has to pick.

\section{Experimental Setup}

\subsection{Datasets}

We conduct our experiment on the publicly available Medical-Diff-VQA dataset \cite{hu2023medical, hu2023expert}. The question-answer pairs in this dataset are derived from the free-text radiology reports from MIMIC-CXR \cite{Johnson2019,johnson2019mimic,goldberger2000physiobank}. The dataset construction followed three main steps: (1) The authors collected keywords by applying ScispaCy \cite{neumann2019scispacy} to extract entities from the reports, which were manually inspected to ensure quality. (2) They then constructed an intermediate KeyInfo dataset through an Extract-Check-Fix cycle, using regular expressions to identify abnormality keywords, manual and automated checks to ensure accuracy, and refining the extraction process until minimal errors were detected, resulting in a dataset that included study details, positive and negative findings, and their attributes. (3) The question-answer pairs were generated based on the entities and attributes from the KeyInfo dataset, and then categorised into seven types:

\begin{enumerate}
    \item \textbf{abnormality} (\textit{``what abnormalities are seen in the $<$location$>$?''})
    \item \textbf{location} (\textit{``Where in the image is the $<$abnormality$>$ located?''})
    \item \textbf{type} (\textit{``What type is the $<$abnormality$>$?''})
    \item \textbf{level} (\textit{``What level is the $<$abnormality$>$?''})
    \item \textbf{view} (\textit{``Which view is this image taken?''})
    \item \textbf{presence} (\textit{``Is there any evidence of $<$abnormality$>$?''})
    \item \textbf{difference} (\textit{``What has changed in the $<$location$>$ area?''})
\end{enumerate}

The \textit{difference} questions, which we refer to as diff-VQA, ask about differences in appearance between the current and a prior scan. In accordance with previous works using this dataset \cite{hu2023expert, cho2024pretraining}, we classify closed questions as those with answers limited to ``yes'' or ``no'', treating them as multiple choice as detailed in Section \ref{sec:vqa_model}. The remaining questions are considered open-ended, with free-form answers.

The Medical-Diff-VQA dataset contains a total of 700,703 question-answer pairs (QA) related to 109,923 pairs of current and prior CXRs. We use the official split; the number of QA pairs and CXR pairs for each data split are presented in Table \ref{tab:vqa_diff_split}. The dataset is divided into training, validation, and testing sets in an 8:1:1 ratio at the study level, ensuring that studies from the same patient appear in only one split to prevent data contamination. To ensure the availability of a second image for differential comparison, only patients from the MIMIC-CXR dataset with more than one prior radiology visit were included.

We use the MIMIC-CXR dataset to train the RG model and the Chest ImaGenome annotation \cite{wu2021chest} to extract anatomical tokens, following the data split indicated in Medical-Diff-VQA at all stages. 

\begin{table}
    \centering
    \begin{tabular}{|c|c|c|}
    \hline
        \textbf{Split} & \textbf{QA pairs} &  \textbf{CXR pairs} \\
    \hline
        Training & 560,563 &  88,098\\
        Validation & 70,070 & 10,864 \\
        Test & 70,070 & 10,963 \\
    \hline
    \end{tabular}
    \caption{Number of QA pairs and CXR pairs for each data split (training/validation/test) in the Medical-Diff-VQA dataset \cite{hu2023medical, hu2023expert}.}
    \label{tab:vqa_diff_split}

\end{table}

\subsection{Implementation Details}

The Report Generator is initialised with random weights and is trained end-to-end for 100 epochs using a cross-entropy loss and Adam optimiser \cite{kingma2014adam}. We set the initial learning rate to $1\times10^{-4}$ and reduce it every 10 epochs by a factor of 0.8. The RG is trained to predict both the finding and impression sections as detailed in Section \ref{sec:rep_gen}. The best-performing model is selected based on the highest BLEU-4 score computed across both the finding and impression sections of the validation set.

The Answer Generator is initialised using the RG weights and is fine-tuned for 100 epochs using the same loss, optimiser and learning rate as the RG. We select the best model based on the highest BLEU-4 score computed across all questions of the validation set.

Each experiment is repeated three times using different random seeds and we report the average in our results.

\subsection{Metrics}

We adopt different metrics based on the type of question, in line with previous studies \cite{hu2023expert, cho2024pretraining}. For ``difference'' type questions, we report natural language generation metrics including BLEU \cite{papineni2002bleu}, METEOR \cite{banerjee2005meteor}, ROUGE \cite{lin2004rouge}, and CIDEr \cite{vedantam2015cider}. We calculate exact-match accuracy for other types of questions, differentiating between open-ended and closed (yes/no) questions.

\subsection{Baselines}

We evaluate our method against existing approaches designed for comparing multiple input images.

For the diff-VQA task, we compare with two methods originally developed for general image-difference captioning: MCCFormers \cite{Qiu_2021_ICCV} and IDCPCL \cite{yao2022image}. Additionally, we benchmark against EKAID \cite{hu2023expert} and PLURAL \cite{cho2024pretraining}, which were previously assessed on the Medical-Diff-VQA dataset.

For all other question types, we compare with state-of-the-art medical VQA methods also evaluated on Medical-Diff-VQA, including MMQ \cite{do2021multiple}, EKAID, and PLURAL.

\section{Results}

\begin{table*}[th!]
\centering
\resizebox{0.85\textwidth}{!}{
    \begin{tabular}{|c|ccccccc|}
    \hline
        Model                            & BLEU-1 & BLEU-2 & BLEU-3 & BLEU-4 & METEOR & ROUGE-L & CIDEr \\
    \hline
        MCCFormers \cite{Qiu_2021_ICCV}  &  0.214 & 0.190  & 0.170  & 0.153  & 0.319  & 0.340   & 0     \\
        IDCPCL \cite{yao2022image}       &  0.614 & 0.541  & 0.474  & 0.414  & 0.303  & 0.582   & 0.703 \\
        EKAID \cite{hu2023expert}        &  0.628 & 0.553  & 0.491  & 0.434  & 0.339  & 0.577   & 1.027\\
        PLURAL \cite{cho2024pretraining} &  0.704 & 0.633  & 0.575  & 0.520  & 0.381  & 0.653   & 1.832\\
    \hline
        AG (w/o report)                  & 0.678  & 0.619 & 0.569 & 0.525 & 0.372 & 0.659 & 2.102 \\
        RG-AG (w/ report)                &  \textbf{0.711} & \textbf{0.650}  & \textbf{0.600}  & \textbf{0.551}  & \textbf{0.384}  & \textbf{0.668}   & \textbf{2.198} \\
    \hline
    \end{tabular}}
    \caption{Comparison results between our proposed approach, both with the report generation step (RG-AG) and without it (AG), and previous methods on the \textit{difference} questions of the MIMIC-diff-VQA dataset \cite{hu2023expert}. We show the \textbf{best results in bold}. All the results of the comparison methods are taken from \cite{cho2024pretraining}.}
    \label{tab:vqa_diff}
\end{table*}

\begin{table}[th!]
\centering
\resizebox{\linewidth}{!}{
    \begin{tabular}{|c|ccc|}
    \hline
        Model                            &  Open Question & Closed Question & All Questions \\
    \hline
        MMQ \cite{do2021multiple}        &  0.115 & 0.108  & 0.115  \\
        EKAID \cite{hu2023expert}        &  0.264 & 0.799  & 0.525  \\
        PLURAL \cite{cho2024pretraining} &  0.512 & \textbf{0.873}  & 0.688  \\
    \hline
        AG (w/o report)                 &  0.509 & 0.865 & 0.683 \\
        RG-AG (w/ report)               & \textbf{0.523} & 0.871 & \textbf{0.693}       \\
    \hline
    \end{tabular}}
    \caption{Comparison results between our proposed approach, both with the report generation step (RG-AG) and without it (AG), and previous methods on all but the \textit{difference} questions of MIMIC-diff-VQA dataset \cite{hu2023expert}. We compute the accuracy (exact match) on the open-ended and the closed-ended questions (yes/no). We show the \textbf{best results in bold}. All the results of the comparison methods are taken from \cite{cho2024pretraining}.}
    \label{tab:vqa_nondiff}
\end{table}

\subsection{VQA Results: Difference \& Non-Difference}

\begin{table*}[th]
    \centering
    \resizebox{0.85\textwidth}{!}{
    \begin{tabular}{|c|c|ccccc|c|}
    \hline
         \textbf{Visual} & \textbf{Text} & BLEU-1 & BLEU-4 & METEOR & ROUGE-L & CIDEr & Acc \\
         \hline
         C         & - & 0.686 & 0.520 & 0.373 & 0.634 & 1.854 & 0.668 \\
         C + P & - & 0.678 & 0.525 & 0.372 & 0.659 & 2.102 & 0.683\\
         C + P & I & 0.679 & 0.523 & 0.369 & 0.654 & 2.111 & 0.690\\
         C + P & F    & 0.690 & 0.533 & 0.374 & 0.655 & 2.117 & 0.691 \\
         -  & F + I & 0.633 & 0.479 & 0.344 & 0.595 & 1.734 & 0.630 \\
         \hline
         C + P & F + I & \textbf{0.711} & \textbf{0.551} & \textbf{0.385} & \textbf{0.668} & \textbf{2.198} & \textbf{0.693} \\
         \hline
         \hline
         \textit{C + P} & \textit{F + I (Ground Truth)} & \textit{0.723} & \textit{0.570} & \textit{0.398} & \textit{0.685} & \textit{2.484} & \textit{0.751}\\
         \hline
    \end{tabular}
    }
    \caption{Ablation results. We test various visual inputs to the Answer Generation (AG) model: the current scan only (C), both current and prior scans (C + P), and no scan (-). Additionally, we test different textual inputs provided alongside the question: the findings section (F), the impression section (I), both sections combined (F + I), and no additional input text (-). In the final row, we present results when the AG model is given the ground truth findings and impression sections. This serves as an upper bound for performance, excluded from direct comparison. We show the \textbf{best results in bold}.}
    \label{tab:vqa_abl}
\end{table*}

We present the diff-VQA results in Table \ref{tab:vqa_diff} and the non-diff-VQA results in Table \ref{tab:vqa_nondiff}, comparing our method with and without the report generation step (RG-AG and AG, respectively) as well as with other state-of-the-art approaches. Our RG-AG method achieves state-of-the-art performance, demonstrating superior results on all NLG metrics for \textit{difference} type questions and improved overall accuracy on the remaining questions. These results suggest that using the pre-trained model on report generation not only to initialise the VQA model -- as done in the PLURAL \cite{cho2024pretraining} and AG methods -- but also to predict the reports and use them to ground the answer generation step helps generate more precise answers, especially for \textit{difference} type questions.  For other question types, the RG-AG model consistently outperforms AG across all metrics. Furthermore, compared to PLURAL, our RG-AG method achieves higher accuracy for open-ended questions but lower accuracy for closed-ended questions (yes/no). This suggests that the intermediate report generation step has a more pronounced impact on open-ended questions, while its influence on closed-ended questions is comparatively limited.

\subsection{Ablation Study}

In this ablation study, we explore the impact of various input components provided to the AG model. The quantitative results, which are detailed in Table \ref{tab:vqa_abl}, illustrate the positive effect of grounding the AG model with relevant sections of radiology reports. By breaking down the components individually, we observe that the Finding section (F) has the most substantial impact on the model's performance when compared to the Impression section (I). This outcome is as expected, as the Finding section typically contains more granular and detailed information about the CXR, offering richer data for the model to process. However, the study also reveals that when both sections are provided in conjunction (F + I), the VQA results show improvement beyond what is achieved with either section alone. This synergy suggests that the combined information from both sections provides a more comprehensive context to the AG model, enhancing its predictions.

Additionally, we examined the scenario where the AG model is provided exclusively with textual information from the report, omitting any visual input from the CXR, to determine whether the generated report alone suffices for accurate answer prediction. The results, however, indicate a noticeable decrement in performance across all metrics, underscoring that the textual reports alone are insufficient. This can be attributed to several factors: predicted reports may contain inaccuracies, or may not contain the answer. These issues highlight that the visual data from the CXR is still necessary for the AG model to generate accurate and reliable answers.

To investigate the impact of inaccuracies, we repeated the experiment using the original expert radiology reports from the MIMIC-CXR dataset in place of the predicted reports. As demonstrated in Table \ref{tab:vqa_abl}, providing the AG model with high-quality curated reports led to improvements across all metrics. This finding underscores the pivotal role of report quality on VQA.

\begin{figure}[t!]
    \centering
    \includegraphics[width=\linewidth]{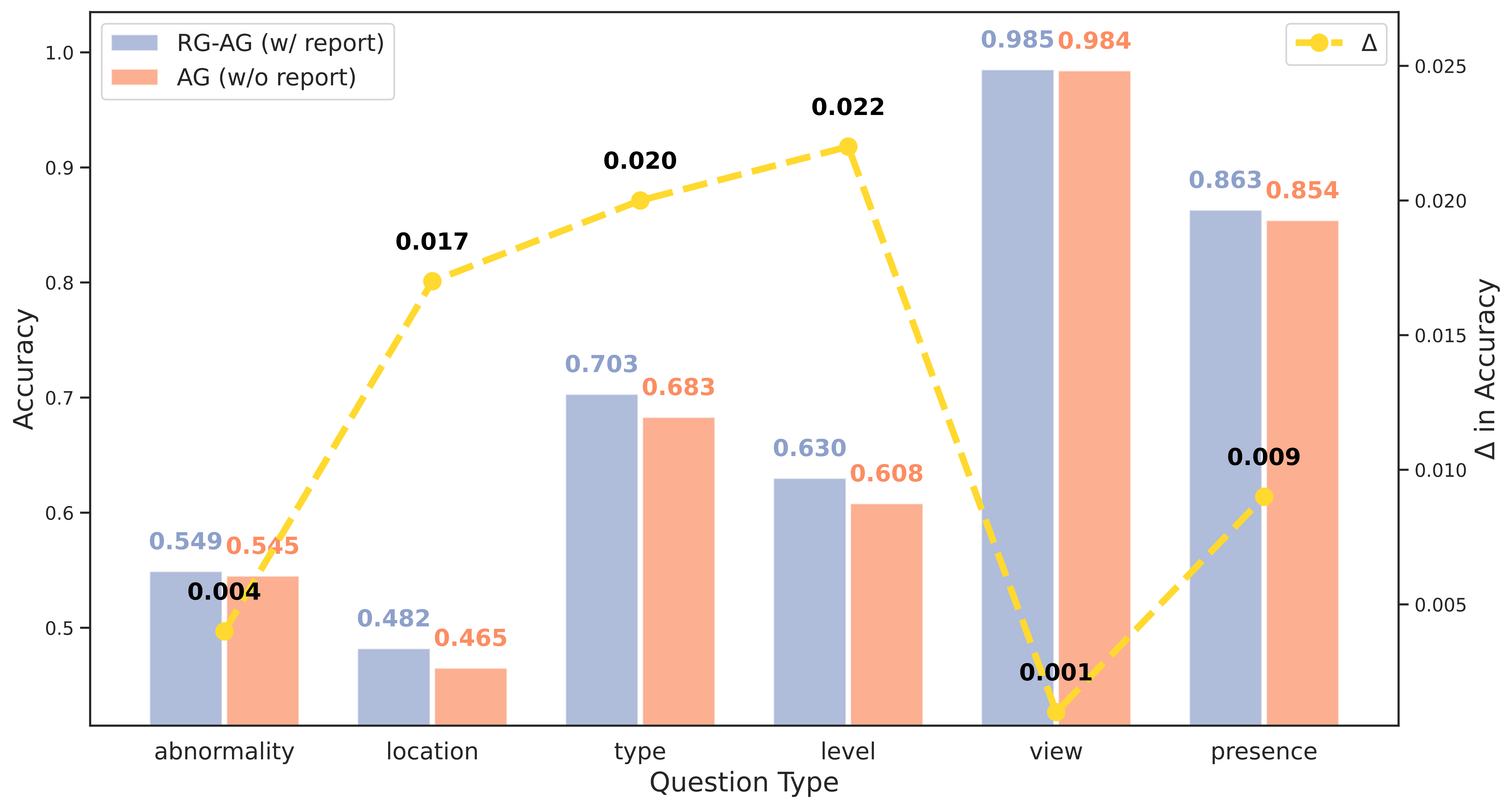}
    \caption{We compare the accuracy of our proposed \textit{RG-AG} model with the baseline \textit{AG} model (which does not include the predicted CXR radiology report as input) for each question type, except for the \textit{difference} questions. We highlight the difference in accuracy ($\Delta$) for each question type.}
    \label{fig:vqa_delta_acc}
\end{figure}

In Figure \ref{fig:vqa_delta_acc}, we present the accuracy across different question types, excluding \textit{difference} questions. Our proposed \textit{RG-AG} model is compared with the baseline \textit{AG} model, which does not incorporate the predicted CXR radiology report as input. The results indicate that not all question types are affected equally when we provide the model with the predicted reports as additional context. Notably, questions related to \textit{location}, \textit{type}, and \textit{level} show the greatest improvement from this additional input.

\begin{figure*}[th!]
    \centering
    \includegraphics[width=0.85\textwidth]{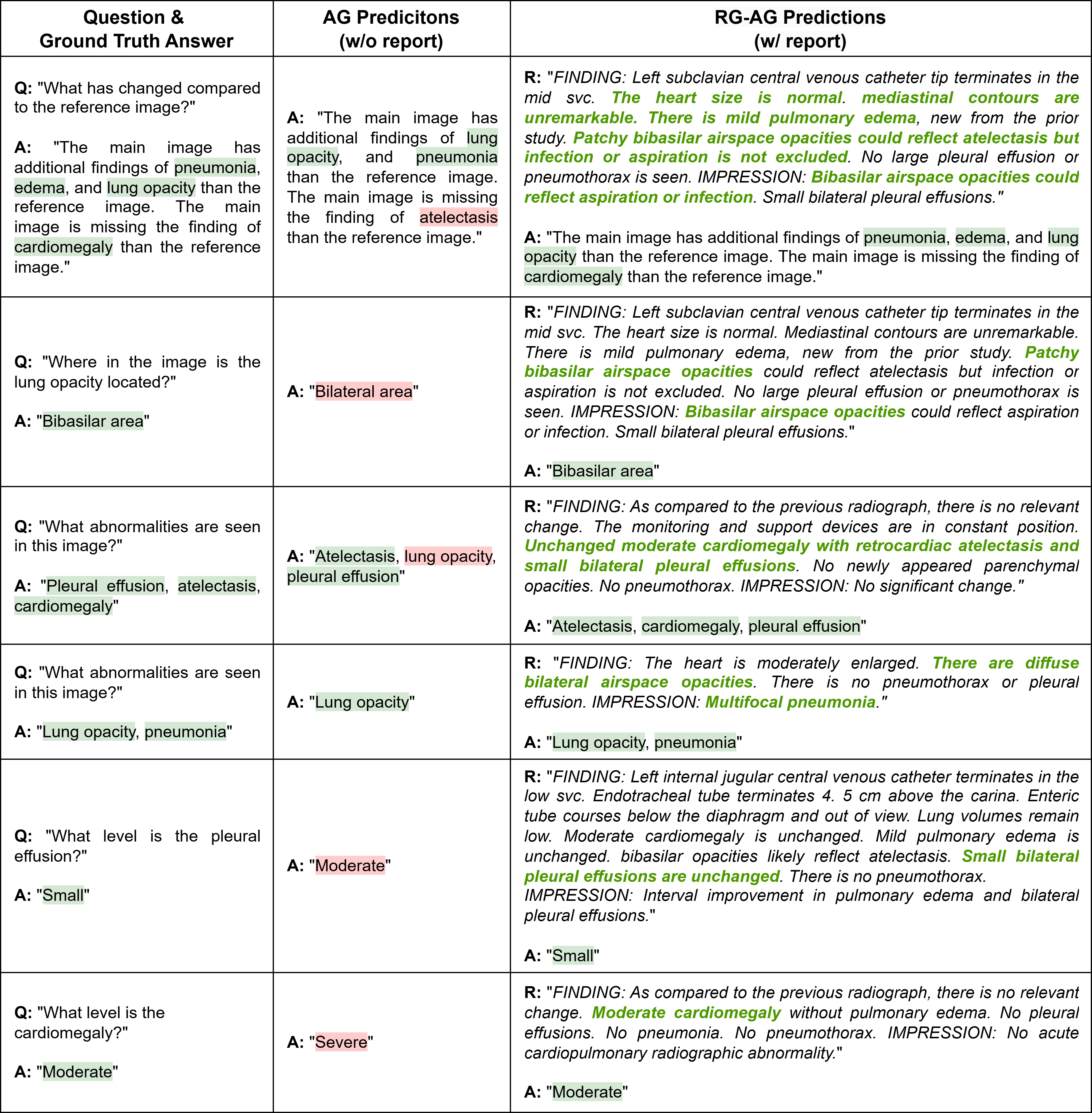}
    \caption{We compare the quality of our predicted answers without the predicted CXR radiology report (\textit{AG} model) and with it (our \textit{RG-AG} model). For each question (Q), we highlight the correct parts of the answer (A) in green and the errors in red. Similarly, in the predicted radiology reports (R), segments containing correct information relevant to the question are shown in green.}
    \label{fig:vqa_qual_fullycorrect}
\end{figure*}

\begin{figure*}[th!]
    \centering
    \includegraphics[width=0.85\textwidth]{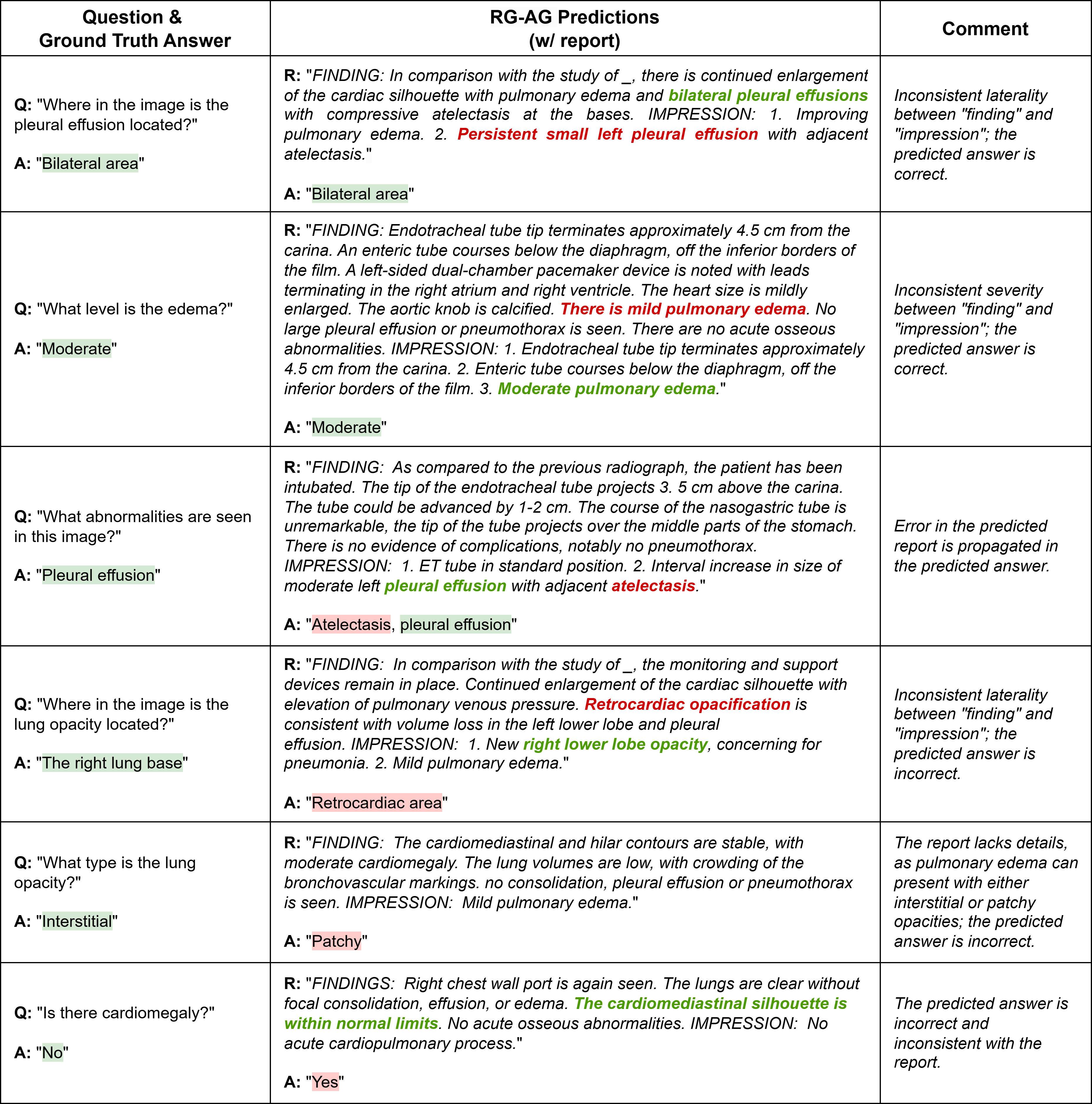}
    \caption{We present borderline and failure cases of our \textit{RG-AG} model, with explanatory comments in the rightmost column to describe the associated errors. For each question (Q), we highlight the correct parts of the answer (A) in green and the errors in red. Similarly, in the predicted radiology reports (R), segments containing correct information relevant to the question are shown in green, and segments inconsistent with the ground truth answer are shown in red.}
    \label{fig:vqa_qual_wrong}
\end{figure*}

\subsection{Qualitative Results}
Finally, we present qualitative results in Figures \ref{fig:vqa_qual_fullycorrect} and  \ref{fig:vqa_qual_wrong} to provide a more comprehensive understanding of the model's behaviour. Figure \ref{fig:vqa_qual_fullycorrect} compares the outputs of the baseline \textit{AG} model---trained without the predicted CXR radiology report---with those of our proposed \textit{RG-AG} model. In these results, we highlight the segments in the predicted report that correctly contain the information needed to predict the correct answer. These results underscore the significance of radiology reports in providing additional evidence for VQA, further validating the importance of this input in enhancing model performance.

Figure \ref{fig:vqa_qual_wrong} highlights several borderline and failure cases of our proposed \textit{RG-AG} model. These examples illustrate: (rows 1–2) the model's robustness to inconsistencies within the predicted reports (\textit{i.e.}, different descriptions between the ``finding'' and ``impression'' sections); (row 3) error propagation from the radiology report to the predicted answer; (row 4) report inconsistency leading to incorrect answer; (row 5) uninformative report results in incorrect answer; and (row 6) inconsistency between the predicted report and the predicted answer.

\section{Limitations}

The Medical-Diff-VQA dataset has limitations related to the origin and scope of the question-answer pairs. Pairs were derived semi-automatically from the ground truth radiology reports, therefore, the style and content of information contained within these reports are highly relevant for answering the questions. However, other types of questions (those not explicitly addressed in the reports) might limit or negate the benefits of grounding the VQA system with the predicted reports. Further, the questions are limited in scope; there are a total of only 18 unique question templates into which relevant keywords are inserted from data-mined lists of 29 disease keywords and approximately 100 keywords describing spatial distribution. This means that more nuanced variation is lacking; for instance for the `difference' question type, rather than asking ``\emph{What has changed in $<$location$>$ area?}'' a clinician might be interested to know ``\emph{Has the effusion resolved as expected?}'' or ``\emph{Is the rate of change of the nodule concerning for a malignant lesion?}''. These questions would require more sophisticated parsing and answer expressivity from any VQA system. Additional evaluation on more varied questions would complement the evaluation presented in this paper.

In terms of methodology, our two-stage RG-AG approach is prone to error propagation between the first and second stage of our pipeline, \textit{i.e.} the predicted reports provided to the AG model might contain errors which can lead to the generation of wrong answers. Furthermore, we generated the \textit{findings} and \textit{impression} sections of the report separately, following the methodology proposed in \cite{cho2024pretraining}. However, this approach can result in inconsistencies between the two sections, potentially leading to incorrect answer predictions (\textit{e.g.}, discrepancies in the laterality or severity of a finding described in each section). A more effective strategy might involve generating the \textit{impression} section based on the predicted \textit{findings}, as is done in the task of summarising radiology reports \cite{delbrouck-etal-2023-overview}.

\section{Conclusion}

We have explored the use of radiology reports as additional context to ground the answer generation task of a CXR VQA system. Our RG-AG method shows state-of-the-art results on the Medical-Diff-VQA dataset, with the most notable improvements noted on the \textit{difference} types of questions compared to prior methods.

We have further investigated the role of the radiology report for VQA in the ablation study, which shows how our RG-AG method achieves the highest overall performance when combining the Finding and Impression sections predicted from a CXR. Furthermore, our results highlight how the quality of the radiology report plays an important role---shown by using the original reports written by expert radiologists---and how providing the AG with visual clues from the CXR is still necessary.

In this paper, we have only addressed whether using the predicted radiology reports can enhance the VQA performance. However, other types of clinical information related to a patient could be provided as evidence to the VQA model to improve its answer generation capability. This strongly depends on the type of questions we want our VQA model to be able to respond to. 

Moreover, we have investigated the problem of grounding the AG with radiology reports, adopting a two-step approach, following the approach proposed in \cite{zhang2024multimodal} for multimodal CoT reasoning. We hypothesise that using a two-stage approach, with each stage implemented using a different model, may only be necessary for smaller models like ours. Larger and more capable models may be able to perform all tasks using a single model, as shown in text-only question-answering \cite{wangself, zhou2023least}.

We leave these two directions as open questions for future work in this space.

\printbibliography
\end{document}